\documentclass[11pt,a4paper]{article}
\usepackage[hyperref]{acl2019}
\usepackage{times}
\usepackage{latexsym}
\usepackage{makecell}
\usepackage{subcaption}
\usepackage{graphicx}
\usepackage{amsfonts}
\usepackage{amsmath}
\usepackage{bm}
\usepackage{mathtools}
\usepackage{url}

\newcolumntype{?}[1]{!{\vrule width #1}}

\aclfinalcopy 
\def\aclpaperid{70} 

\newcommand\blfootnote[1]{%
  \begingroup
  \renewcommand\thefootnote{}\footnote{#1}%
  \addtocounter{footnote}{-1}%
  \endgroup
}

\title{Building a Production Model for Retrieval-Based Chatbots}

\author{Kyle Swanson\textsuperscript{1\textdagger}, Lili Yu\textsuperscript{2}, Christopher Fox\textsuperscript{2}, Jeremy Wohlwend\textsuperscript{2}, Tao Lei\textsuperscript{2} \\
\textsuperscript{1}Massachusetts Institute of Technology \\
\textsuperscript{2}ASAPP, Inc. \\
{\tt swansonk@mit.edu} \\
{\tt \{liliyu, cdfox, jeremy, tao\}@asapp.com}}

\begin{document}
\maketitle
\begin{abstract}
Response suggestion is an important task for building human-computer conversation systems. Recent approaches to conversation modeling have introduced new model architectures with impressive results, but relatively little attention has been paid to whether these models would be practical in a production setting. In this paper, we describe the unique challenges of building a production retrieval-based conversation system, which selects outputs from a whitelist of candidate responses. To address these challenges, we propose a dual encoder architecture which performs rapid inference and scales well with the size of the whitelist. We also introduce and compare two methods for generating whitelists, and we carry out a comprehensive analysis of the model and whitelists. Experimental results on a large, proprietary help desk chat dataset, including both offline metrics and a human evaluation, indicate production-quality performance and illustrate key lessons about conversation modeling in practice.
\end{abstract}

\section{Introduction}
\label{sec:introduction}

\blfootnote{\textsuperscript{\textdagger}Work done primarily while an intern at ASAPP, Inc.}
Predicting a response given conversational context is a critical task for building open-domain chatbots and dialogue systems.
Recently developed conversational systems typically use either a generative or a retrieval approach for producing responses~\cite{wang2013,ji2014,vinyals2015,serban2015,li2016,xing2016,deb2019diversifying}. 
While both of these approaches have demonstrated strong performance in the literature, retrieval methods often enjoy better control over response quality than generative approaches.
In particular, such methods select outputs from a \textit{whitelist} of candidate responses, which can be pre-screened and revised for desired qualities such as sentence fluency and diversity.

Most previous work on retrieval models has concentrated on designing neural architectures to improve response selection.
For instance, several works have improved model performance by encoding multi-turn conversation context instead of single-turn context~\cite{serban2015,zhou2016,wu2017}.
More recent efforts~\cite{zhou2018,zhang2018} have explored using more advanced architectures, such as the Transformer~\cite{vaswani2017}, to better learn the mapping between the context and the candidate responses.

Relatively little effort, however, has been devoted to the practical considerations of using such models in a real-world production setting. 
For example, one critical consideration rarely discussed in the literature is the inference speed of the deployed model.
While recent methods introduce rich computation, such as cross-attention \cite{zhou2018}, to improve the modeling between the conversational context and candidate response, the model outputs must be re-computed for every pair of context and response.
As a consequence, these models are not well-suited to a production setting where the size of the response whitelist can easily extend into the thousands. 

Another critical concern is the whitelist selection process and the associated retrieval evaluation.
Most prior work have reported Recall@$k$ on a small set of randomly selected responses which include the true response sent by the agent ~\cite{lowe2015,zhou2016,zhou2018,wu2017,zhang2018}.
However, this over-simplified evaluation may not provide a useful indication of performance in production, where the whitelist is not randomly selected, is significantly larger, and may not contain the target response.

In this paper, we explore and evaluate model and whitelist design choices for building retrieval-based conversation systems in production. 
We present a dual encoder architecture that is optimized to select among as many as 10,000 responses within a couple tens of milliseconds.
The model makes use of a fast recurrent network implementation~\cite{lei2018} and multi-headed attention~\cite{lin2017} and achieves over a 4.1x inference speedup over traditional encoders such as LSTM~\cite{hochretier1997}.
The independent dual encoding allows pre-computing the embeddings of candidate responses, thereby making the approach highly scalable with the size of the whitelist.
In addition, we compare two approaches for generating the response candidates, and we conduct a comprehensive analysis of our model and whitelists on a large, real-world help desk dataset, using human evaluation and metrics that are more relevant to use in a production setting.

\section{Related Work}
\label{sec:related_work}

This paper extends the line of work on conversational retrieval models for multi-turn response selection \cite{lowe2015,alrfou2016,zhou2016,zhou2018,wu2016,wu2017,yan2016,lu2017,zhang2018,shalyminov2018,deb2019diversifying,yang2019hybrid}. Our model is most similar to \citet{lowe2015}, who construct the context of the conversation by concatenating all previous utterances. They use an RNN to separately encode the context and each candidate response, and they then compute a matching score between the context and response representations to determine the best response for that context.

Other recent work has explored more complex methods of incorporating information from the context of a conversation. \citet{serban2015} and \citet{zhou2016} employ a hierarchical architecture in which they encode the context using RNNs at both the word level and the utterance level. In contrast to these models, which generate a single context encoding, \citet{wu2017} designed a network that matches a response to each utterance in the context individually.

While many of the models cited above implement their RNNs with an LSTM \cite{hochretier1997}, we instead use an SRU \cite{lei2018}. SRU uses light recurrence, which makes it highly parallelizable, and \citet{lei2018} showed that it trains 5-9x faster than cuDNN LSTM. SRU also exhibits a significant speedup in inference time compared to LSTM (by a factor of 4.1x in our experiments), which is particularly relevant in a production setting. Furthermore, \citet{lei2018} showed that SRU matches or exceeds the performance of models using LSTMs or the Transformer architecture \cite{vaswani2017} on a number of NLP tasks, meaning significant speed gains can be achieved without a drop in performance.

Despite the abundance of prior work on retrieval models for dialogue, whitelist selection has received relatively little attention. Since practical use of conversational models has typically not been addressed, most models are evaluated on their ability to select the correct response from a small list of randomly sampled responses \cite{lowe2015}. Another option, from \citet{wu2017}, is to use \textit{Apache Lucene}\footnote{\url{http://lucene.apache.org/}} to select a list of response candidates relevant to each context. However, neither method produces a single whitelist that can be used for every context and reviewed for quality. The closest work to ours is \citet{lu2017}, who build a whitelist using a $k$-means clustering of responses. We extend this work by doing a more comprehensive analysis of different whitelist selection methods, and we further analyze the effect of whitelist size on performance.

\section{Model Architecture}
\label{sec:model_architecture}

\begin{figure*}
\centering
\begin{subfigure}{.45\textwidth}
  \centering
  \includegraphics[width=\linewidth]{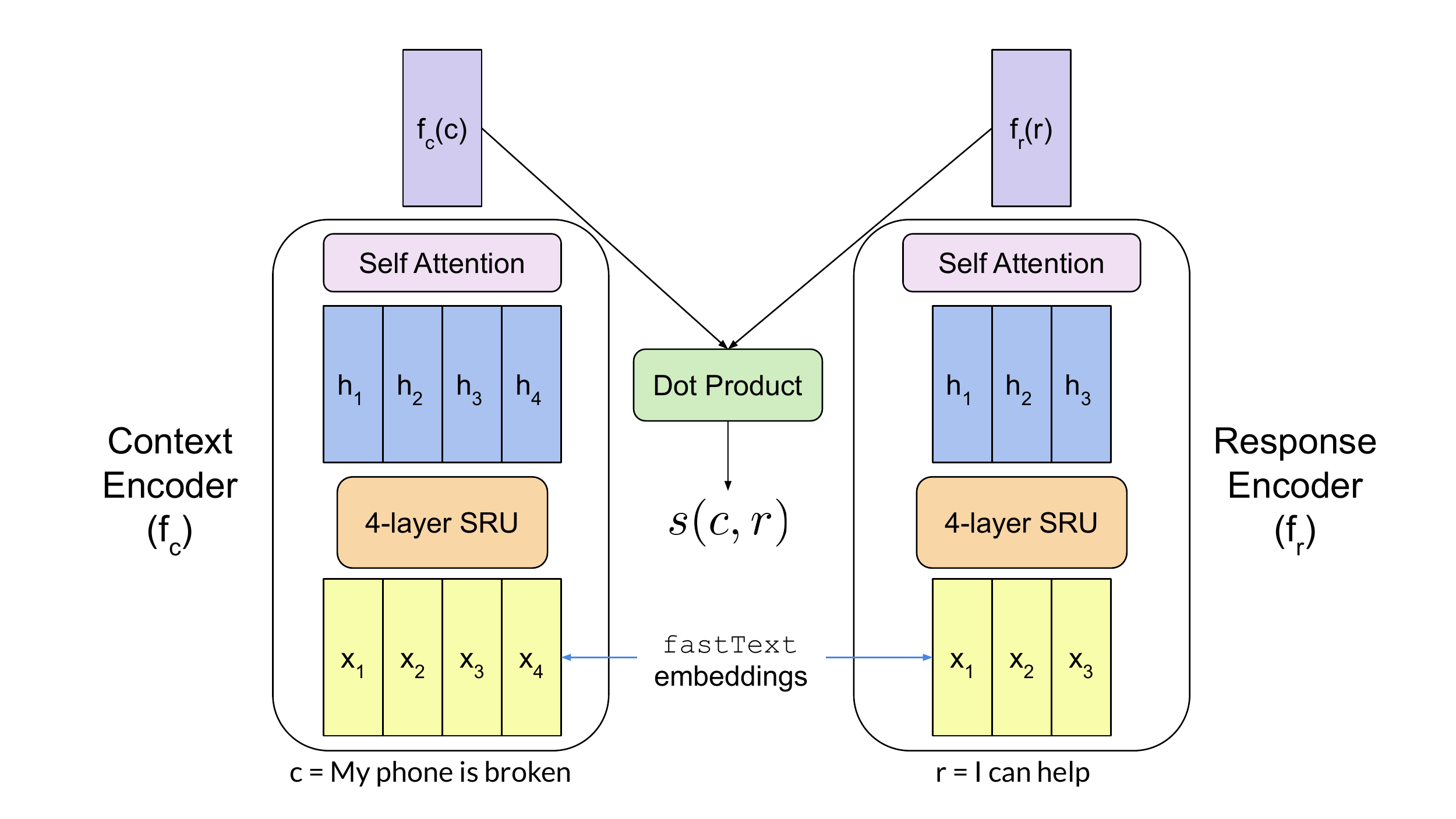}
\end{subfigure}
\begin{subfigure}{.45\textwidth}
  \centering
  \includegraphics[width=\linewidth]{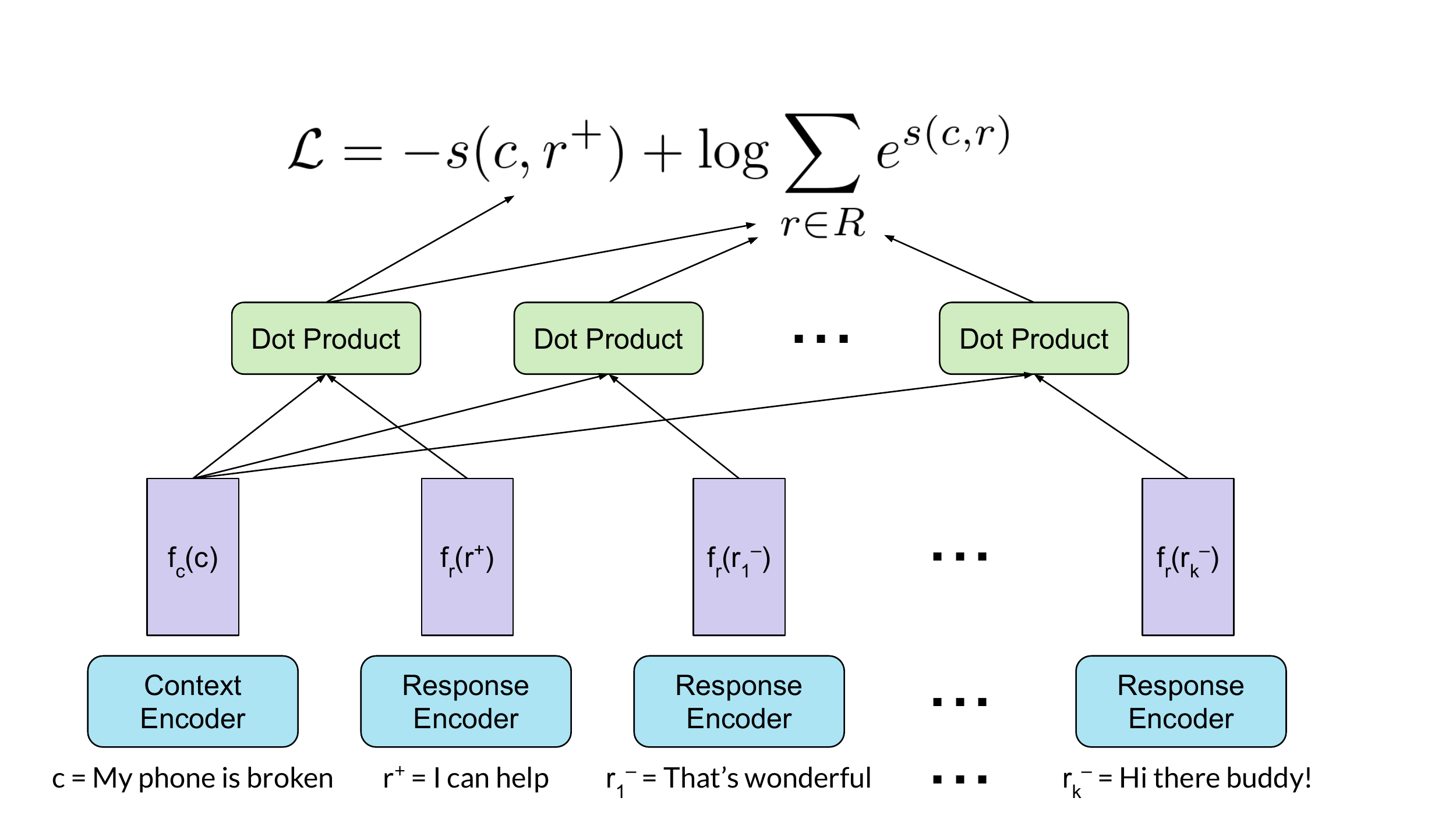}
\end{subfigure}
\caption{(Left) The dual encoder architecture, which takes as input a context $c$ and a response $r$ and computes the score $s(c,r)$. (Right) Computing the model's loss based on the scores between a context $c$ and each response in $R = \{ r^+, r_1^-, \dots, r_k^- \}$, where $r^+$ is the actual agent response and  $r_1^-, \dots, r_k^-$ are randomly sampled agent responses.}
\label{fig:model}
\end{figure*}

Next we describe the architecture of our retrieval model. The two inputs to the model are a context $c$, which is a concatenation of all utterances in the conversation, and a candidate response $r$. In the context, we use special tokens to indicate whether each utterance comes from the customer or the agent. The model outputs a score $s(c,r)$ indicating the relevance of the response to the context. The model architecture is described in detail below and is illustrated in Figure \ref{fig:model}.

\subsection{Dual Encoders}
\label{ssec:dual_encoders}

At the core of our model are two neural encoders $f_c$ and $f_r$ to encode the context and the response, respectively. These encoders have identical architectures but learn separate weights. 

Each encoder takes a sequence of tokens $w = \{w_1, w_2, \dots, w_n\}$ as input, which is either a context or a response.
Due to the prevalence of typos in both user and agent utterances in real chats, we use {\tt fastText} \cite{bojanowski2016} as the word embedding method. {\tt fastText} learns both word-level and character-level features and is therefore more robust to misspellings. We pretrained {\tt fastText}\footnote{\url{https://github.com/facebookresearch/fastText}} embeddings on a corpus of 15M utterances from help desk conversations and then fixed the embeddings while training the neural encoders.

Each encoder consists of a recurrent neural network followed by a multi-headed attention layer~\cite{lin2017} to perform pooling. 
We use multi-layer, bidirectional SRUs as the recurrent network. 
Each layer of the SRU involves the following computation:
\begin{equation}
    \begin{split}
        \textbf{f}_t &= \sigma(\textbf{W}_f \textbf{x}_t + \textbf{v}_f \odot \textbf{c}_{t-1} + \textbf{b}_f) \\
        \textbf{c}_t &= \textbf{f}_t \odot \textbf{c}_{t-1} + (1 - \textbf{f}_t) \odot (\textbf{W} \textbf{x}_t) \\
        \textbf{r}_t &= \sigma(\textbf{W}_r \textbf{x}_t + \textbf{v}_r \odot \textbf{c}_{t-1} + \textbf{b}_r) \\
        \textbf{h}_t &= \textbf{r}_t \odot \textbf{c}_t + (1 - \textbf{r}_t) \odot \textbf{x}_t
    \end{split}
\end{equation}
where $\sigma$ is the sigmoid activation function, $\textbf{W}, \textbf{W}_f, \textbf{W}_r \in \mathbb{R}^{d_h \times d_e}$ are learned parameter matrices, and $\textbf{v}_f, \textbf{v}_r, \textbf{b}_f, \textbf{b}_v \in \mathbb{R}^{d_h}$ are learned parameter vectors.

The multi-headed attention layer compresses the encoded sequence $\textbf{h} = \{\textbf{h}_1, \textbf{h}_2, \dots, \textbf{h}_n\}$ into a single vector. 
For each attention head $i$, attention weights are generating with the following computation:
\begin{equation}
    \bm{\alpha}^{(i)} = {\tt softmax}(\sigma(\textbf{h}^T \textbf{W}^{(i)}_a) \textbf{v}^{(i)}_a)
\end{equation}
where $\sigma$ is a non-linear activation function, $\textbf{W}^{(i)}_a \in \mathbb{R}^{d_h \times d_a}$ is a learned parameter matrix, and $\textbf{v}^{(i)}_a \in \mathbb{R}^{d_a}$ is a learned parameter vector.

The encoded sequence representation is then pooled to a single vector for each attention head $i$ by summing the attended representations:
\begin{equation}
    \widetilde{\textbf{h}}^{(i)} = \sum_{j=1}^n \bm{\alpha}^{(i)}_j \textbf{h}_j\ .
\end{equation}

Finally, the pooled encodings are averaged across the $n_h$ attention heads:
\begin{equation}
    \widetilde{\textbf{h}} = \frac{1}{n_h} \sum_{i=1}^{n_h} \widetilde{\textbf{h}}^{(i)}\ .
\end{equation}
The output of the encoder is the vector $f(w) = \widetilde{\textbf{h}}$.

\subsection{Scoring}
\label{ssec:scoring}

To determine the relevance of a response $r$ to a context $c$, our model computes a matching score between the context encoding $f_c(c)$ and the response encoding $f_r(r)$. This score is simply the dot product of the encodings:
\begin{equation}
    s(c, r) = f_c(c) \cdot f_r(r)\ .
\end{equation}

\subsection{Training}
\label{ssec:training}

We optimize the model to maximize the score between the context $c$ and the response $r^+$ actually sent by the agent while minimizing the score between the context and each of $k$ random (``negative'') responses $r_1^-, \dots, r_k^-$. This is accomplished by training the model to minimize the cross-entropy loss:
\begin{equation}
    \mathcal{L} = -s(c, r^+) + \log \sum_{r \in R} e^{s(c, r)}
    \label{eq:loss}
\end{equation}
where $R = \{r^+, r_1^-, \dots, r_k^-\}$.

Although negative responses could be sampled separately for each context-response pair, we instead use a method inspired by \citet{logeswaran2018} and share a set of negative responses across all examples in a batch. Specifically, for each batch, we sample $k$ responses from the set of all agent responses (weighted according to response frequency), and we use those $k$ responses as the negative responses for every context-response pair in the batch. This has the benefit of reducing the number of responses that need to be encoded in each batch of size $b$ from $\mathcal{O}(bk)$ to $\mathcal{O}(b + k)$, thereby significantly accelerating training.

\begin{table*}[!h]
\centering
\begin{tabular}{|c|c|}
    \hline
    Conversations & 594,555 \\
    \hline
    Utterances & 15,217,773 \\
    \hline
    Customer utterances & 6,943,940 \\
    \hline
    Agent utterances & 8,273,833 \\
    \hline
    Mean conversation length (\# utterances) & 25.60 \\
    \hline
    Mean utterance length (\# tokens) & 12.70 \\
    \hline
    Mean customer utterance length (\# tokens) & 7.53 \\
    \hline
    Mean agent utterance length (\# tokens) & 17.15 \\
    \hline
\end{tabular}
\caption{Summary statistics for the propriety help desk dataset.}
\label{tab:data_stats}
\end{table*}

\subsection{Whitelist Generation}
\label{ssec:whitelist_generation}

After training, we experimented with two methods of creating the whitelist from which our model selects responses at inference time. For each method, we created both a 1,000 response whitelist and a 10,000 response whitelist. Having a whitelist with any more than 10,000 responses would likely make a manual review infeasible.

\paragraph{Frequency-Based Method.}

Responses that are sent frequently are more likely to be relevant in multiple conversations and are less likely to contain errors. Therefore, one method of building a high-quality whitelist is simply to collect messages that are sent often. We created frequency-based whitelists by selecting the 1,000 or 10,000 most common agent responses, after accounting for minor variations in capitalization, punctuation, and whitespace.

\paragraph{Clustering-Based Method.}

Although selecting responses based on frequency may help guarantee quality, manual examination of the frequency whitelists showed that they contained many redundant responses. Therefore, we experimented with a clustering-based whitelist selection method in the hope of reducing redundancy and increasing response diversity. Specifically, we encoded all agent responses using our response encoder $f_r$ and then used $k$-means clustering with $k=1,000$ or $k=10,000$ to cluster the responses. We then selected the most common response from each cluster to create the whitelists.

\section{Experiments and Results}
\label{sec:experiments_and_results}

We evaluated our model and whitelists on a large, proprietary help desk chat dataset using several offline metrics and a human evaluation. We particularly emphasize metrics relevant to production, such as inference speed and Recall@$k$ from a large candidate set. The human evaluation illustrates how our model and whitelists compare to each other and to the responses sent by a real human agent.

\subsection{Data}
\label{ssec:data}

\begin{table}
\centering
\begin{tabular}{|p{7.5cm}|}
    \hline
    \multicolumn{1}{|c|}{\textbf{Example Conversation}} \\
    \Xhline{3\arrayrulewidth}
    \textbf{Customer:} I would like to pay my bill can you help me \\
    \textbf{Agent:} I can definitely help you to pay your bill. Are we going to work with the account logged in now? \\
    \textbf{Customer:} Yes it still says there is no money on my account \\
    \textbf{Agent:} I understand that. I have reviewed your account and its shows here that the payment has been posted and you're all good until next month service. \\
    \textbf{Customer:} Oh ok thank you for all your help \\
    \textbf{Agent:} You're welcome. Anything for a valued customer like you! \\
    \hline
\end{tabular}
\caption{A sample conversation from the propriety help desk chat dataset. The sample has been lightly edited to remove proprietary information.}
\label{tab:chat_example}
\end{table}

The help desk chat dataset used in our experiments consists of 15M utterances from 595K conversations. We randomly split the conversations into train, validation, and test sets with 80\%, 10\%, and 10\% of the conversations, respectively. Since each conversation includes several agent responses, each of which produces a context-response example, our dataset consists of 6.6M training examples, 828K validation examples, and 828K test examples. Additional dataset statistics are provided in Table \ref{tab:data_stats}. An example chat conversation can be seen in Table \ref{tab:chat_example}.

\subsection{Model Details}
\label{ssec:model_details}

We implemented the dual encoder model using PyTorch~\cite{paszke2017}. 
We use pre-trained {\tt fastText} embeddings of dimension $d_e=300$, a 4-layer bidirectional SRU\footnote{SRU code available at \url{https://github.com/taolei87/sru/tree/master/sru}} with hidden size $d_h=300$, and multi-headed attention with 16 heads and a hidden size of $d_a=64$. The batch size was 200 and we used $k=200$ negative responses for each positive response. To ensure quick encoding even for long inputs, contexts were restricted to the 500 most recent tokens and responses were restricted to the 100 most recent tokens\footnote{A context with 500 tokens contains 39 utterances on average, which is typically more than enough to understand the topic of conversation. Almost all responses are shorter than 100 tokens.}. The model was optimized using Adam~\cite{kingma2014} with the Noam learning rate schedule from ~\citet{vaswani2017}. The model was trained for 30 epochs, with each epoch limited to 10,000 training batches (2M training examples). Training took about 32 hours on a single Tesla V100 GPU.

\subsection{Results and Analysis}
\label{ssec:performance_and_analysis}

\paragraph{AUC and AUC@$p$.}

To determine the model's ability to use context to distinguish between true responses and negative responses, we use the metrics AUC and AUC@$p$. AUC is the area under the receiver operating characteristic curve when using the score $s(c,r)$ to determine whether each response is the true response or a negative response. AUC@$p$ is the area under the portion of the ROC curve where the false positive rate is $\leq p$, renormalized so that the maximum AUC@$p$ is 1.

The performance of our model according to these AUC metrics can be seen in Table~\ref{tab:auc}. The high AUC indicates that our model can easily distinguish between the true response and negative responses. Furthermore, the AUC@$p$ numbers show that the model has a relatively high true positive rate even under the difficult requirement of a low false positive rate.

\begin{table}
\centering
\begin{tabular}{|c|c|c|}
    \hline
    \textbf{Metric} & \textbf{Validation} & \textbf{Test} \\
    \Xhline{3\arrayrulewidth}
    AUC & 0.991 & 0.977 \\
    AUC@0.1 & 0.925 & 0.885 \\
    AUC@0.05 & 0.871 & 0.816 \\
    AUC@0.01 & 0.677 & 0.630 \\
    \hline
\end{tabular}
\caption{AUC and AUC@$p$ of our model on the propriety help desk dataset.}
\label{tab:auc}
\end{table}

\begin{table}
\centering
\begin{tabular}{|c|c|c|c|c|}
    \hline
    \textbf{Candidates} & \textbf{R@1} & \textbf{R@3} & \textbf{R@5} & \textbf{R@10} \\
    \Xhline{3\arrayrulewidth}
    10 & 0.892 & 0.979 & 0.987 & 1 \\
    100 & 0.686 & 0.842 & 0.894 & 0.948 \\
    1,000 & 0.449 & 0.611 & 0.677 & 0.760 \\
    10,000 & 0.234 & 0.360 & 0.421 & 0.505 \\
    \hline
\end{tabular}
\caption{Recall@$k$ from $n$ response candidates for different values of $n$ using random whitelists. Each random whitelist includes the correct response along with $n-1$ randomly selected responses.}
\label{tab:random_recall}
\end{table}

\paragraph{Recall and Whitelist Size.}

In order to determine our model's ability to select the correct response from a whitelist, we use recall at $k$ from $n$ (R\textsubscript{$n$}@$k$), which is the proportion of times that the true response is ranked as one of the top $k$ responses in a whitelist containing $n$ candidate responses.

Table \ref{tab:random_recall} shows R\textsubscript{$n$}@$k$ on the test set for different values of $n$ and $k$ when using a random whitelist, meaning a whitelist which contains the true response and $n-1$ randomly sampled responses\footnote{To be precise, we sampled responses without replacement weighted according to the frequency with which the response was sent by agents.}. As discussed in the introduction, most prior work evaluate their models using a random whitelist with $n=10$ candidates. However, a production whitelist needs to contain hundreds or thousands of response candidates in order to provide relevant responses in a variety of contexts. Therefore, a more meaningful metric for production purposes is R\textsubscript{$n$}@$k$ for $n \geq 100$. Table \ref{tab:random_recall} shows that recall drops significantly as $n$ grows, meaning that the R\textsubscript{10}@$k$ evaluation performed by prior work may significantly overstate model performance in a production setting.

\begin{table}
\centering
\resizebox{0.49\textwidth}{!}{
    \begin{tabular}{|c|c|c|c|c|c|}
        \hline
        \textbf{Whitelist} & \textbf{R@1} & \textbf{R@3} & \textbf{R@5} & \textbf{R@10} & \textbf{BLEU} \\
        \Xhline{3\arrayrulewidth}
        Random 10K+ & 0.252 & 0.400 & 0.472 & 0.560 &  37.71 \\
        Frequency 10K+ & 0.257 & 0.389 & 0.455 & 0.544 & 41.34 \\
        Clustering 10K+ & 0.230 & 0.376 & 0.447 & 0.541 &  37.59\\
        \hline
        Random 1K+ & 0.496 & 0.663 & 0.728 & 0.805 & 59.28\\
        Frequency 1K+ & 0.513 & 0.666 & 0.726 & 0.794 & 67.05\\
        Clustering 1K+ & 0.481 & 0.667 & 0.745 & 0.835 & 61.88 \\
        \hline
        \hline
        Frequency 10K & 0.136 & 0.261 & 0.327 & 0.420 & 30.46\\
        Clustering 10K & 0.164 & 0.292 & 0.360 & 0.457 &  31.47\\
        \hline
        Frequency 1K & 0.273 & 0.465 & 0.550 & 0.658 &  47.13 \\
        Clustering 1K & 0.331 & 0.542 & 0.650 & 0.782 & 49.26\\
        \hline
    \end{tabular}
}
\caption{Recall@$k$ for random, frequency, and clustering whitelists of different sizes. The ``+'' indicates that the true response is added to the whitelist.}
\label{tab:random_vs_frequency_vs_clustering}
\end{table}

\begin{table}
\centering
\begin{tabular}{|c|c|c|c|c|}
    \hline
    \textbf{Whitelist} & \textbf{R@1} & \textbf{Coverage} \\
    \Xhline{3\arrayrulewidth}
    Frequency 10K & 0.136 & 45.04\% \\
    Clustering 10K & 0.164 & 38.38\% \\
    \hline
    Frequency 1K & 0.273 & 33.38\% \\
    Clustering 1K & 0.331 & 23.28\% \\
    \hline
\end{tabular}
\caption{Recall@1 versus coverage for frequency and clustering whitelists.}
\label{tab:frequency_vs_clustering}
\end{table}

\paragraph{Comparison Between Whitelists.}
An interesting question we would like to address is whether a random whitelist serves as a good proxy for whitelists generated using other methods. 
To this end, we also evaluate recall on the frequency and clustering whitelists from Section \ref{ssec:whitelist_generation}.

First, we compute recall when the true response is added to the whitelist, as in the case of the random whitelists described above. Second, we compute recall only on the subset of examples for which the true response is already contained in the whitelist. The latter recall measure is more relevant to a production setting since the true response is not known at inference time and therefore cannot be artificially added to the whitelist.

The results in Table \ref{tab:random_vs_frequency_vs_clustering} show that the three types of whitelists perform comparably to each other when the true response is added.
However, in the more realistic second case, when recall is only computed on examples with a response already in the whitelist, performance on the frequency and clustering whitelists drops significantly.

Additionally, we compute the BLEU ~\cite{papineni2002bleu, ward2002corpus} scores between the true responses and the best suggested responses. The BLEU score allows us to measure the semantic similarity when the true and suggested responses are not exactly matched. The BLEU scores computed with the frequency and clustering whitelists are slightly higher than those computed with random whitelists. 

\paragraph{Recall versus Coverage.}

Although recall is a good measure of performance, recall alone is not a sufficient criterion for whitelist selection. The recall results in Table \ref{tab:random_vs_frequency_vs_clustering} seem to indicate that the clustering-based whitelists are strictly superior to the frequency-based whitelists in the realistic case when we only consider responses that are already contained in the whitelist, but this analysis fails to account for the frequency with which this is the case. For instance, a whitelist may have very high recall but may only include responses that were sent infrequently by agents, meaning the whitelist will perform well for a handful of conversations but will be irrelevant in most other cases.

To quantify this effect, we introduce the notion of \textit{coverage}, which is the percent of all context-response pairs where the agent response appears in the whitelist, after accounting for minor deviations in capitalization, punctuation, and whitespace. A whitelist that contains responses that are sent more frequently by agents will therefore have a higher coverage.

Table \ref{tab:frequency_vs_clustering} shows R@1 and coverage for the frequency and clustering whitelists. While the clustering whitelists have higher recall, the frequency whitelists have higher coverage. This is to be expected since the frequency whitelists were specifically chosen to maximize the frequency of the included responses. Since both recall and coverage are necessary to provide good responses for a wide range of conversations, these results indicate the importance of considering the trade-off between recall and coverage inherent in a given whitelist selection method.

It may be interesting in future work to further investigate these trade-offs in order to identify a whitelist selection method that can simultaneously optimize recall and coverage.

\begin{table}
\centering
\resizebox{0.49\textwidth}{!}{
    \begin{tabular}{|c|c|c|c?{0.5mm}c|}
        \hline
        \textbf{Whitelist} & \textbf{Great} & \textbf{Good} & \textbf{Bad} & \textbf{Accept} \\
        \Xhline{3\arrayrulewidth}
        Freq. 1K & 54\% & 26\% & 20\% & 80\% \\
        Cluster. 1K & 55\% & 21\% & 23\% & 77\% \\
        Freq. 10K & 56\% & 24\% & 21\% & 80\% \\
        Cluster. 10K & 57\% & 23\% & 20\% & 80\% \\
        \hline
        Real response & 60\% & 24\% & 16\% & 84\% \\
        \hline
    \end{tabular}
}
\caption{Results of the human evaluation of the responses produced by our model. A response is acceptable if it is either good or great. Note: Numbers may not add up to 100\% due to rounding.}
\label{tab:human}
\end{table}

\paragraph{Human Evaluation.}
While offline metrics are indicative of model performance, the best measure of performance is a human evaluation of the model's predictions. Therefore, we performed a small-scale human evaluation of our model and whitelists. We selected 322 contexts from the test set and used our model to generate responses from the Frequency 10K, Frequency 1K, Clustering 10K, and Clustering 1K whitelists. Three human annotators were shown each context followed by five responses: one from each of the four whitelists and the true response sent by the agent. The annotators were blinded to the source of each response. The annotators were asked to rate each response according to the following categories:

\vspace{3mm}

\noindent
\textbf{Bad:} The response is not relevant to the context.

\noindent
\textbf{Good:} The response is relevant to the context but is vague or generic.

\noindent
\textbf{Great:} The response is relevant to the context and directly addresses the issue at hand.

\vspace{3mm}

\noindent
For example, three such responses for the context ``My phone is broken'' would be:

\vspace{3mm}

\noindent
\textbf{Bad response:} Goodbye!

\noindent
\textbf{Good response:} I'm sorry to hear that.

\noindent
\textbf{Great response:} I'm sorry to hear that your phone is broken.

\vspace{3mm}

The results of the human evaluation are in Table \ref{tab:human}. Our proposed system works well, selecting acceptable (i.e. good or great) responses about 80\% of the time and selecting great responses more than 50\% of the time.

Interestingly, the size and type of whitelist seem to have little effect on performance, indicating that all the whitelists contain responses appropriate to a variety of conversational contexts. Since the frequency whitelists are simpler to generate than the clustering whitelists and since the 1K whitelists contain fewer responses to manually review than the 10K whitelists, the Frequency 1K whitelist is the preferred whitelist for our production system.

\begin{table}
\centering
\begin{tabular}{|c|c|c|c|}
    \hline
    \textbf{Encoder} & \textbf{Layer} & \textbf{Params} & \textbf{Time} \\
    \Xhline{3\arrayrulewidth}
    SRU & 2 & 3.7M & 14.7 \\
    SRU & 4 & 8.0M & 21.9 \\
    LSTM & 2 & 7.3M & 90.9 \\
    LSTM & 4 & 15.9M & 174.8  \\
    \hline
    $ $+rank response& - & - & 0.9 \\
    \hline
\end{tabular}

\caption{Inference time (milliseconds) of our model to encode a context using an SRU or an LSTM encoder on a single CPU core. The last row shows the extra time needed to compare the response encoding to 10,000 cached candidate response encodings in order to find the best response.}
\label{tab:inference_speed}
\end{table}
\paragraph{Inference Speed.}
A major constraint in a production system is the speed with which the system can respond to users.
To demonstrate the benefit of using an SRU encoder instead of an LSTM encoder in production, we compared the speed with which they encode a random conversation context at inference time, averaged over 1,000 samples.
We used a single core of Intel Core i9 2.9 GHz CPU. 
As seen in Table \ref{tab:inference_speed}, an SRU encoder is over 4x faster than an LSTM encoder with a similar number of parameters, making it more suitable for production use.

Table~\ref{tab:inference_speed} also highlights the scalability of using a dual encoder architecture.
Since the embeddings of the candidate responses are independent from the conversation context, the embeddings of the whitelist responses can be pre-computed and stored as a matrix. 
Retrieving the best candidate once the context is encoded takes a negligible amount of time compared to the time to encode the context.

\begin{table*}[!t]
\centering
\begin{tabular}{|c|c|c|c|c|}
    \hline
    \textbf{Model} & \textbf{Parameters} & \textbf{Validation AUC@0.05} & \textbf{Test AUC@0.05} \\
    \Xhline{2\arrayrulewidth}
    Base & 8.0M & \textbf{0.871} & 0.816 \\
    \hline
    4L SRU $\rightarrow$ 2L LSTM & 7.3M & 0.864 & \textbf{0.829} \\
    \hline
    4L SRU $\rightarrow$ 2L SRU & 7.8M & 0.856 & \textbf{0.829} \\
    \hline
    Flat $\rightarrow$ hierarchical & 12.4M & 0.825 & 0.559 \\
    \hline
    Cross entropy $\rightarrow$ hinge loss & 8.0M & 0.765 & 0.693 \\
    \hline
    6.6M $\rightarrow$ 1M examples & 8.0M & 0.835 & 0.694 \\
    \hline
    6.6M $\rightarrow$ 100K examples & 8.0M & 0.565 & 0.417 \\
    \hline
    200 $\rightarrow$ 100 negatives & 8.0M & 0.864 & 0.647 \\
    \hline
    200 $\rightarrow$ 10 negatives & 8.0M & 0.720 & 0.412 \\
    \hline
\end{tabular}
\caption{An ablation study showing the effect of different model architectures and training regimes on performance on the proprietary help desk dataset.}
\label{tab:ablation}
\end{table*}

\paragraph{Ablation analysis.}
Finally, we performed an ablation analysis to identify the effect of different aspects of the model architecture and training regime. The results are shown in Table \ref{tab:ablation}, and details of the model variants are available in the Appendix.

As Table \ref{tab:ablation} shows, the training set size and the number of negative responses for each positive response are the most important factors in model performance. The model performs significantly worse when trained with hinge loss instead of cross-entropy loss, indicating the importance of the loss function. We also experimented with a hierarchical encoder, where two different recurrent neural networks are used to encode contexts, one at the word level and one at the utterance level. We observed no advantage to using a hierachical encoder, despite its complexity and popularity for encoding conversations \cite{serban2015,zhou2016}. Finally, we see that a 2 layer LSTM performs similarly to either a 4 layer or a 2 layer SRU with a comparable number of parameters. Since the SRU is more than 4x faster at inference time with the same level of performance, it is the preferred encoder architecture.

\section{Conclusion}
\label{sec:conclusion}

In this paper, we present a fast dual encoder neural model for retrieval-based human-computer conversations. 
We address technical considerations specific to the production setting, and we evaluate our model and two whitelist generation methods on a large help desk chat dataset. We observe that traditional offline evaluation metrics significantly overestimate model performance, indicating the importance of using evaluation metrics more relevant to a production setting. Furthermore, we find that our proposed model performs well, both on offline metrics and on a human evaluation. Due to its strong performance and its speed at inference time, we conclude that our proposed model is suitable for use in a production conversational system.

One important direction for future work is a deeper analysis of the whitelist selection process. Although our analysis found similar performance across whitelists according to a human evaluation, our offline metrics indicate underlying trade-offs between different characteristics of the whitelists such as recall and coverage. A better understanding the implications of these trade-offs may lead to improved whitelist generation methods, thereby further improving the performance of retrieval-based models.

\section*{Acknowledgments}
\label{sec:acknowledgments}

We would like to thank Howard Chen for the invaluable conversations we had with him during the development of our model. We would also like to thank Anna Folinsky and the ASAPP annotation team for their help performing the human evaluation, and Hugh Perkins for his support on the experimental environment setup. Thank you as well to Ethan Elenberg, Kevin Yang, and Adam Yala for reviewing early drafts of this paper and providing valuable feedback. Finally, thank you to the anonymous reviewers for their constructive feedback.

\bibliography{acl2019}

\begin{thebibliography}{28}
\expandafter\ifx\csname natexlab\endcsname\relax\def\natexlab#1{#1}\fi

\bibitem[{Al-Rfou et~al.(2016)Al-Rfou, Pickett, Snaider, hsuan Sung, Strope,
  and Kurzweil}]{alrfou2016}
Rami Al-Rfou, Marc Pickett, Javier Snaider, Yun hsuan Sung, Brian Strope, and
  Ray Kurzweil. 2016.
\newblock \href {https://arxiv.org/abs/1606.00372} {Conversational contextual
  cues: The case of personalization and history for response ranking}.
\newblock \emph{arXiv preprint arXiv:1606.00372}.

\bibitem[{Bojanowski et~al.(2016)Bojanowski, Grave, Joulin, and
  Mikolov}]{bojanowski2016}
Piotr Bojanowski, Edouard Grave, Armand Joulin, and Tomas Mikolov. 2016.
\newblock \href {https://arxiv.org/abs/1607.04606} {Enriching word vectors with
  subword information}.
\newblock \emph{arXiv preprint arXiv:1607.04606}.

\bibitem[{Deb et~al.(2019)Deb, Bailey, and Shokouhi}]{deb2019diversifying}
Budhaditya Deb, Peter Bailey, and Milad Shokouhi. 2019.
\newblock \href {https://arxiv.org/abs/1903.10630} {Diversifying reply
  suggestions using a matching-conditional variational autoencoder}.
\newblock \emph{arXiv preprint arXiv:1903.10630}.

\bibitem[{Hochreiter and Schmidhuber(1997)}]{hochretier1997}
Sepp Hochreiter and J{\"u}rgen Schmidhuber. 1997.
\newblock Long short-term memory.
\newblock \emph{Neural computation}, 9(8):1735--1780.

\bibitem[{Ji et~al.(2014)Ji, Lu, and Li}]{ji2014}
Zongcheng Ji, Zhengdong Lu, and Hang Li. 2014.
\newblock \href {https://arxiv.org/abs/1408.6988} {An information retrieval
  approach to short text conversation}.
\newblock \emph{arXiv preprint arXiv:1408.6988}.

\bibitem[{Kingma and Ba(2014)}]{kingma2014}
Diederik~P. Kingma and Jimmy Ba. 2014.
\newblock \href {https://arxiv.org/abs/1412.6980} {Adam: A method for
  stochastic optimization}.
\newblock \emph{arXiv preprint arXiv:1412.6980}.

\bibitem[{Lei et~al.(2018)Lei, Zhang, Wang, Dai, and Artzi}]{lei2018}
Tao Lei, Yu~Zhang, Sida~I. Wang, Hui Dai, and Yoav Artzi. 2018.
\newblock \href {https://arxiv.org/abs/1709.02755} {Simple recurrent units for
  highly parallelizable recurrence}.
\newblock \emph{arXiv preprint arXiv:1709.02755}.

\bibitem[{Li et~al.(2016)Li, Galley, Brockett, Spithourakis, Gao, and
  Dolan}]{li2016}
Jiwei Li, Michel Galley, Chris Brockett, Georgios~P. Spithourakis, Jianfeng
  Gao, and Bill Dolan. 2016.
\newblock \href {https://arxiv.org/abs/1603.06155} {A persona-based neural
  conversation model}.
\newblock \emph{arXiv preprint arXiv:1603.06155}.

\bibitem[{Lin et~al.(2017)Lin, Feng, dos Santos, Yu, Xiang, Zhou, and
  Bengio}]{lin2017}
Zhouhan Lin, Minwei Feng, Cicero~Nogueira dos Santos, Mo~Yu, Bing Xiang, Bowen
  Zhou, and Yoshua Bengio. 2017.
\newblock \href {https://arxiv.org/abs/1703.03130} {A structured self-attentive
  sentence embedding}.
\newblock \emph{arXiv preprint arXiv:1703.03130}.

\bibitem[{Logeswaran and Lee(2018)}]{logeswaran2018}
Lajanugen Logeswaran and Honglak Lee. 2018.
\newblock \href {https://arxiv.org/abs/1803.02893} {An efficient framework for
  learning sentence representations}.
\newblock \emph{arXiv preprint arXiv:1803.02893}.

\bibitem[{Lowe et~al.(2015)Lowe, Pow, Serban, and Pineau}]{lowe2015}
Ryan Lowe, Nissan Pow, Iulian Serban, and Joelle Pineau. 2015.
\newblock \href {https://arxiv.org/abs/1506.08909} {The ubuntu dialogue corpus:
  A large dataset for research in unstructured multi-turn dialogue systems}.

\bibitem[{Lu et~al.(2017)Lu, Keung, Zhang, Sun, and Bhardwaj}]{lu2017}
Yichao Lu, Phillip Keung, Shaonan Zhang, Jason Sun, and Vikas Bhardwaj. 2017.
\newblock \href {https://arxiv.org/abs/1703.09439} {A practical approach to
  dialogue response generation in closed domains}.
\newblock \emph{arXiv preprint arXiv:1703.09439}.

\bibitem[{Papineni et~al.(2002)Papineni, Roukos, Ward, and
  Zhu}]{papineni2002bleu}
Kishore Papineni, Salim Roukos, Todd Ward, and Wei-Jing Zhu. 2002.
\newblock Bleu: a method for automatic evaluation of machine translation.
\newblock In \emph{Proceedings of the 40th annual meeting on association for
  computational linguistics}, pages 311--318. Association for Computational
  Linguistics.

\bibitem[{Paszke et~al.(2017)Paszke, Gross, Chintala, Chanan, Yang, DeVito,
  Lin, Desmaison, Antiga, and Lerer}]{paszke2017}
Adam Paszke, Sam Gross, Soumith Chintala, Gregory Chanan, Edward Yang, Zachary
  DeVito, Zeming Lin, Alban Desmaison, Luca Antiga, and Adam Lerer. 2017.
\newblock \href {https://openreview.net/forum?id=BJJsrmfCZ} {Automatic
  differentiation in pytorch}.
\newblock In \emph{NIPS-W}.

\bibitem[{Serban et~al.(2015)Serban, Sordoni, Bengio, Courville, and
  Pineau}]{serban2015}
Iulian~V. Serban, Alessandro Sordoni, Yoshua Bengio, Aaron Courville, and
  Joelle Pineau. 2015.
\newblock \href {https://arxiv.org/abs/1507.04808} {Building end-to-end
  dialogue systems using generative hierarchical neural network models}.
\newblock \emph{arXiv preprint arXiv:1507.04808}.

\bibitem[{Shalyminov et~al.(2018)Shalyminov, Dušek, and
  Lemon}]{shalyminov2018}
Igor Shalyminov, Ondřej Dušek, and Oliver Lemon. 2018.
\newblock \href {https://arxiv.org/abs/1811.00967} {Neural response ranking for
  social conversation: A data-efficient approach}.
\newblock \emph{arXiv preprint arXiv:1811.00967}.

\bibitem[{Vaswani et~al.(2017)Vaswani, Shazeer, Parmar, Uszkoreit, Jones,
  Gomez, Kaiser, and Polosukhin}]{vaswani2017}
Ashish Vaswani, Noam Shazeer, Niki Parmar, Jakob Uszkoreit, Llion Jones,
  Aidan~N. Gomez, Lukasz Kaiser, and Illia Polosukhin. 2017.
\newblock \href {https://arxiv.org/abs/1706.03762} {Attention is all you need}.
\newblock \emph{arXiv preprint arXiv:1706.03762}.

\bibitem[{Vinyals and Le(2015)}]{vinyals2015}
Oriol Vinyals and Quoc Le. 2015.
\newblock \href {https://arxiv.org/abs/1506.05869} {A neural conversational
  model}.
\newblock \emph{arXiv preprint arXiv:1506.05869}.

\bibitem[{Wang et~al.(2013)Wang, Lu, Li, and Chen}]{wang2013}
Hao Wang, Zhengdong Lu, Hang Li, and Enhong Chen. 2013.
\newblock \href {http://www.anthology.aclweb.org/D/D13/D13-1096.pdf} {A dataset
  for research on short-text conversation}.
\newblock \emph{Proceedings of the 2013 Conference on Empirical Methods in
  Natural Language Processing}, pages 935--945.

\bibitem[{Ward and Reeder(2002)}]{ward2002corpus}
Kishore Papineni Salim Roukos~Todd Ward and John Henderson~Florence Reeder.
  2002.
\newblock Corpus-based comprehensive and diagnostic mt evaluation: Initial
  arabic, chinese, french, and spanish results.

\bibitem[{Wu et~al.(2016)Wu, Wang, and Xue}]{wu2016}
Bowen Wu, Baoxun Wang, and Hui Xue. 2016.
\newblock \href {https://www.aclweb.org/anthology/C/C16/C16-1063.pdf} {Ranking
  responses oriented to conversational relevance in chat-bots}.
\newblock \emph{COLING16}.

\bibitem[{Wu et~al.(2017)Wu, Wu, Xing, Zhou, and Li}]{wu2017}
Yu~Wu, Wei Wu, Chen Xing, Ming Zhou, and Zhoujun Li. 2017.
\newblock \href {http://dx.doi.org/10.18653/v1/P17-1046} {Sequential matching
  network: A new architecture for multi-turn response selection in
  retrieval-based chatbots}.
\newblock \emph{Proceedings of the 55th Annual Meeting of the Association for
  Computational Linguistics}.

\bibitem[{Xing et~al.(2016)Xing, Wu, Wu, Liu, Huang, Zhou, and Ma}]{xing2016}
Chen Xing, Wei Wu, Yu~Wu, Jie Liu, Yalou Huang, Ming Zhou, and Wei-Ying Ma.
  2016.
\newblock \href {https://arxiv.org/abs/1606.08340} {Topic aware neural response
  generation}.
\newblock \emph{arXiv preprint arXiv:1606.08340}.

\bibitem[{Yan et~al.(2016)Yan, Song, and Wu}]{yan2016}
Rui Yan, Yiping Song, and Hua Wu. 2016.
\newblock \href {http://dx.doi.org/10.1145/2911451.2911542} {Learning to
  respond with deep neural networks for retrieval-based human-computer
  conversation system}.
\newblock \emph{SIGIR}, pages 55--64.

\bibitem[{Yang et~al.(2019)Yang, Hu, Qiu, Qu, Gao, Croft, Liu, Shen, and
  Liu}]{yang2019hybrid}
Liu Yang, Junjie Hu, Minghui Qiu, Chen Qu, Jianfeng Gao, W.~Bruce Croft,
  Xiaodong Liu, Yelong Shen, and Jingjing Liu. 2019.
\newblock \href {https://arxiv.org/abs/1904.09068} {A hybrid
  retrieval-generation neural conversation model}.
\newblock \emph{arXiv preprint arXiv:1904.09068}.

\bibitem[{Zhang et~al.(2018)Zhang, Li, Zhu, Zhao, and Liu}]{zhang2018}
Zhuosheng Zhang, Jiangtong Li, Pengfei Zhu, Hai Zhao, and Gongshen Liu. 2018.
\newblock \href {https://arxiv.org/abs/1806.09102} {Modeling multi-turn
  conversation with deep utterance aggregation}.
\newblock \emph{arXiv preprint arXiv:1806.09102}.

\bibitem[{Zhou et~al.(2016)Zhou, Dong, Wu, Zhao, Yu, Tian, Liu, and
  Yan}]{zhou2016}
Xiangyang Zhou, Daxiang Dong, Hua Wu, Shiqi Zhao, Dianhai Yu, Hao Tian, Xuan
  Liu, and Rui Yan. 2016.
\newblock \href {http://www.aclweb.org/anthology/D16-1036} {Multi-view response
  selection for human-computer conversation}.
\newblock \emph{Proceedings of the 2016 Conference on Empirical Methods in
  Natural Language Processing}, pages 372--381.

\bibitem[{Zhou et~al.(2018)Zhou, Li, Dong, Liu, Chen, Zhao, Yu, and
  Wu}]{zhou2018}
Xiangyang Zhou, Lu~Li, Daxiang Dong, Yi~Liu, Ying Chen, Wayne~Xin Zhao, Dianhai
  Yu, and Hua Wu. 2018.
\newblock \href {http://www.aclweb.org/anthology/P18-1103} {Multi-turn response
  selection for chatbots with deep attention matching network}.
\newblock \emph{Proceedings of the 56th Annual Meeting of the Association for
  Computational Linguistics}, pages 1118--1127.

\end{thebibliography}
\bibliographystyle{acl_natbib}

\clearpage
\appendix

\section{Appendix}
\label{sec:appendix}

\subsection{Ablation Study}
\label{ssec:ablation_study}

Table \ref{tab:ablation} shows the results of an ablation study we performed to identify the most important components of our model architecture and training regime. Each variant is described below.

\vspace{3mm}

\noindent
\textbf{Base.} This is the model architecture described in Section \ref{sec:model_architecture}.

\vspace{3mm}

\noindent
\textbf{4L SRU $\rightarrow$ 2L LSTM.} We replace the 4 layer SRU encoder with a 2 layer LSTM encoder, which has a comparable number of parameters when using the same hidden sizes.

\vspace{3mm}

\noindent
\textbf{4L SRU $\rightarrow$ 2L SRU.} We use an SRU with 2 layers instead of 4 layers. In order to match parameters, we use a hidden size of $d_h=475$ instead of $d_h=300$ in the model with 2 layers.

\vspace{3mm}

\noindent
\textbf{Flat $\rightarrow$ hierarchical.} We replace the SRU encoder with two SRU encoders, one which operates at the word level and one which operates at the utterance level, following the architectures of \citet{serban2015,wu2017}.

\vspace{3mm}

\noindent
\textbf{Cross entropy $\rightarrow$ hinge loss.} Instead of using the cross-entropy loss defined in Equation \ref{eq:loss}, we use the hinge loss, which is defined as:
\begin{equation}
    \mathcal{L} = \sum_{i=1}^k |s(c,r^+) - s(c,r_i^-) + m|\
\end{equation}
where the margin $m=0.25$ encourages separation between the score of the correct response and the score of each negative response.

\vspace{3mm}

\noindent
\textbf{6.6M $\rightarrow$ 1M examples.} We train on a dataset with 1 million examples instead of the full 6.6 million training examples.

\vspace{3mm}

\noindent
\textbf{6.6M $\rightarrow$ 100K examples.} We trained on a dataset with 100 thousand examples instead of the full 6.6 million training examples.

\vspace{3mm}

\noindent
\textbf{200 $\rightarrow$ 100 negatives.} During training, we sample 100 negatives for each context-response pair instead of 200 negatives.

\vspace{3mm}

\noindent
\textbf{200 $\rightarrow$ 10 negatives.} During training, we sample 10 negatives for each context-response pair instead of 200 negatives.

\end{document}